\documentclass[twoside,11pt]{article}

\usepackage{automl2020}

\usepackage{amsmath}
\usepackage{booktabs}
\usepackage{algorithm}
\usepackage{algorithmic}
\usepackage{makecell}
\usepackage{color}
\usepackage{graphicx} 
\usepackage{caption}
\usepackage{subcaption}
\jmlrheading{1}

\firstpageno{1}

\begin{document}
% \vspace{-0.5cm}
\title{Uncertainty aware
Search Framework for Multi-Objective Bayesian Optimization with
Constraints}
% \vspace{-0.5cm}
\author{\name Syrine Belakaria \email syrine.belakaria@wsu.edu \\
       \addr Washington State University
       \AND
       \name  Aryan Deshwal \email aryan.deshwal@wsu.edu \\
       \addr Washington State University
        \AND
       \name Janardhan Rao Doppa \email jana.doppa@wsu.edu\\
       \addr Washington State University}

\maketitle
% \vspace{-0.5cm}
\begin{abstract}%   <- trailing '%' for backward compatibility of .sty file
We consider the problem of constrained multi-objective (MO) blackbox
optimization using expensive function evaluations, where the
goal is to approximate the true Pareto set of solutions satisfying a set of constraints while minimizing the number of function evaluations. We propose a novel framework named Uncertainty-aware Search framework for Multi-Objective Optimization with
Constraints (USeMOC) to efficiently select the sequence of inputs for evaluation to solve this problem. The selection method of USeMOC consists of solving a cheap constrained MO
optimization problem via surrogate models of the true functions
to identify the most promising candidates and picking the
best candidate based on a measure of uncertainty. We applied this framework to optimize the design of a multi-output switched-capacitor voltage regulator via expensive simulations. 
Our experimental results show that USeMOC is able to achieve more than 90\% reduction in the number of simulations needed to uncover optimized circuits.

\end{abstract}
% \vspace{-0.5cm}
\section{Introduction}
Many engineering and scientific applications involve making design choices to optimize multiple objectives. A representative example includes designing new materials to optimize strength, elasticity, and durability. There are three %common 
challenges in solving these kind of optimization problems: {\bf 1)} The objective functions are unknown and we need to perform expensive experiments to evaluate each candidate design. For example, performing physical lab experiments for material design application. {\bf 2)} The objectives are conflicting in nature and all of them cannot be optimized simultaneously. {\bf 3)} The designs that are part of the solution should satisfy a set of constraints. Therefore, we need to find the {\em Pareto optimal} set of solutions satisfying the constraints. A solution is called Pareto optimal if it cannot be improved in any of the objectives without compromising some other objective. The overall 
goal is to approximate the true Pareto set satisfying the constraints while minimizing the number of function evaluations. 

Bayesian Optimization (BO) (\cite{shahriari2016taking}) is an effective framework to solve blackbox optimization problems with expensive function evaluations. The key idea behind BO is to
build a cheap surrogate model (e.g., Gaussian Process (\cite{williams2006gaussian})) using the real experimental evaluations; and employ it to intelligently select the sequence of function evaluations using an acquisition function, e.g., expected improvement (EI). There is a large body of literature on single-objective BO algorithms (\cite{shahriari2016taking,BOCS,PSR,DBO}) and their applications including hyper-parameter tuning of machine learning methods (\cite{snoek2012practical,kotthoff2017auto}). However, there is relatively less work on the more challenging problem of BO for multiple objectives (\cite{knowles2006parego,emmerich2008computation,hernandez2016predictive,MESMO,USEMO}) and very limited prior methods to address constrained MO problems (\cite{garrido2019predictive,feliot2017bayesian}). PESMOC (\cite{garrido2019predictive}) is a state-of-the-art approach that relies on the principle of input space entropy search. However, it is computationally expensive to optimize the acquisition function behind PESMOC. A series of approximations are performed to improve the efficiency potentially at the expense of accuracy.

In this paper, we propose a novel {\bf U}ncertainty-aware {\bf Se}arch framework for  {\bf M}ultiple {\bf O}bjectives optimizing with {\bf C}onstraints (USeMOC) to overcome the drawbacks of prior methods. The key insight is to improve uncertainty management via a two-stage search procedure to select candidate inputs for evaluation. First, it solves a cheap constrained MO optimization problem defined in terms of the acquisition functions (one for each unknown objective) to identify a list of promising candidates. Second, it selects the best candidate from this list based on a measure of uncertainty. We demonstrate the efficacy of USeMOC over prior methods using a real-world application from the domain of analog circuit design via expensive simulations.

\section{Background and Problem Setup}

\noindent {\bf Bayesian Optimization Framework.} 
Let $\mathcal{X} \subseteq \Re^d$ be an input space. We assume an unknown real-valued objective function $F: \mathcal{X} \mapsto \Re$, which can evaluate each input $x \in \mathcal{X}$ to produce an evaluation $y$ = $F(x)$.  Each evaluation $F(x)$ is expensive in terms of the consumed resources. The main goal is to find an input $x^* \in \mathcal{X}$ that approximately optimizes $F$ via a limited number of function evaluations. BO algorithms learn a cheap surrogate model from training data obtained from past function evaluations. They intelligently select the next input for evaluation by trading-off exploration and exploitation to quickly direct the search towards optimal inputs.
The three key elements of BO framework are:

%\vspace{0.5ex}

 {\bf 1) Statistical Model} of 
 $F(x)$. {\em Gaussian Process (GP)} (\cite{williams2006gaussian}) is the most commonly used model.
 A GP over a space $\mathcal{X}$ is a random process from $\mathcal{X}$ to $\Re$. It is characterized by a mean function $\mu : \mathcal{X} \times \mathcal{X} \mapsto \Re$ and a covariance or kernel function $\kappa$. If a function $F$ is sampled from GP($\mu$, $\kappa$), then $F(x)$ is distributed normally $\mathcal{N}(\mu(x), \kappa(x,x))$ for a finite set of inputs from $x \in \mathcal{X}$.
 
 %\vspace{0.5ex}
 
 {\bf 2) Acquisition Function} (\textsc{Af}) to score the utility of evaluating a candidate input $x \in \mathcal{X}$ based on the statistical model. Some popular acquisition functions include expected improvement (EI), upper confidence bound (UCB) and lower confidence bound (LCB). For the sake of completeness, we formally define the acquisition functions employed in this work noting that any other acquisition function can be employed within USeMOC. 
 %\vspace{-0.2cm}
 \begin{align}
    & UCB(x)=\mu(x)+\beta^{1/2} \sigma(x) \\\label{ucbeq} 
    & LCB(x)=\mu(x)-\beta^{1/2} \sigma(x)\\\label{lcbeq}
    & EI(x)= \sigma(x)(\alpha\Phi(\alpha)+\phi(\alpha)) , ~ \alpha=\frac{\tau-\mu(x)}{\sigma(x)}
\end{align}
where $\mu(x)$ and $\sigma(x)$ correspond to the mean and standard deviation of the prediction from statistical model, $\beta$ is a parameter that balances exploration and exploitation; $\tau$ is the best uncovered input; and $\Phi$ and $\phi$ are the CDF and PDF of normal distribution respectively.

%\vspace{0.5ex}

{\bf 3) Optimization Procedure} to select the best scoring candidate input according to \textsc{Af} via statistical model, e.g., DIRECT (\cite{jones1993lipschitzian}).

\vspace{1.0ex}

\noindent {\bf Multi-Objective Optimization (MOO) Problem with Constraints.}  Without loss of generality, our goal is to minimize $k \geq 2$ real-valued objective functions $F_1(x), F_2(x),\cdots,F_k(x)$ while satisfying $L$ constraints $C_1(x), C_2(x),\cdots,C_L(x)$ over continuous space  $X \subseteq \Re^d$. Each evaluation of an input $x \in \mathcal{X}$ produces a vector of objective values $Y$ = $(y_1, y_2,\cdots,y_k)$ where $y_i$ = $F_i(x)$ for all $i \in \{1,2, \cdots, k\}$ and a vector of constraints values $C$ = $(c_1, c_2,\cdots,c_L)$ where $c_i$ = $C_i(x)$ for all $i \in \{1,2, \cdots, L\}$. Constraints in real world problems usually take one of the following three forms: \textbf{[Type 1]} $C_i$ is a function of the input $x$ and can be expressed declaratively (i.e., white-box constraint); \textbf{[Type 2]} $C_i$ is black-box constraint; and \textbf{[Type 3]} $C_i$ is a function of both the input $x$ and a combination of the black-box objective functions. 
We say that a point $x$ {\em Pareto-dominates} another point $x'$ if $F_i(x) \leq F_i(x') \hspace{1mm} \forall{i}$ and there exists some $j \in \{1, 2, \cdots,k\}$ such that $F_j(x) < F_j(x')$. The optimal solution of constrained MOO problem is a set of points $\mathcal{X}^* \subset \mathcal{X}$ such that no point $x' \in \mathcal{X} \setminus \mathcal{X}^*$ Pareto-dominates a point $x \in \mathcal{X}^*$ and all points in $\mathcal{X}^*$ satisfies the problem constraints. The solution set $\mathcal{X}^*$ is called the Pareto set and the corresponding set of function values is called the Pareto front. Our goal is to approximate $\mathcal{X}^*$ while minimizing the number of function evaluations.
%\vspace{-0.5cm}
\section{Uncertainty-Aware Search Framework}
\label{section4}
%\vspace{-0.2cm}
In this section, we provide the details of USeMOC framework for solving constrained multi-objective optimization problems. First, we provide an overview of USeMOC followed by the details of its two main components.
%\vspace{-0.3cm}
\subsection{Overview of USeMOC Framework} 
USeMOC is an iterative
algorithm that involves four key steps. First, We build statistical models $\mathcal{M}_1, \mathcal{M}_2,\cdots,\mathcal{M}_k$ for each of the $k$ objective functions from the training data in the form of past function evaluations. Second, we select a set of promising candidate inputs $\mathcal{X}_p$ by solving a constrained cheap MO optimization problem defined using the statistical models. Specifically, multiple objectives of the cheap MO problem correspond to $\textsc{Af}(\mathcal{M}_1,x), \textsc{Af}(\mathcal{M}_2,x),\cdots, \textsc{Af}(\mathcal{M}_k,x)$ respectively.  
Any standard acquisition function \textsc{Af} from single-objective BO (e.g., EI, LCB) can be used for this purpose. Additionally, if the constraint is black-box, it is modeled by a GP $\mathcal{M}_{c_i}$ and its predictive mean $\mu_{c_i}(x)$ is used  instead for reasoning. If the value of a constraint $C_i$ depends on the evaluation of one (or more) objectives $F_i(x)$, we employ the predictive mean $\mu_i(x)$. The Pareto set $\mathcal{X}_p$ corresponds to the inputs with different trade-offs in the utility space for $k$ unknown functions satisfying the constraints. Third, we select the best candidate input $x_s \in \mathcal{X}_p$  from the Pareto set that maximizes some form of uncertainty measure for evaluation. Fourth, the selected input $x_s$ is used for evaluation to get the corresponding function evaluations: $y_1$=$F_1(x_s)$, $y_2$=$F_2(x_s)$,$\cdots$,$y_k$=$F_k(x_s)$.
Algorithm~\ref{alg:USeMO} provides the algorithmic pseudocode for USeMOC.

\begin{algorithm}[h]
% \scriptsize
\caption{USeMOC Framework}
\label{alg:USeMO}
\textbf{Input}: $\mathcal{X}$, input space; $F_1(x), F_2(x),\cdots,F_k(x)$, $k$ blackbox objective functions; \textsc{Af}, acquisition function; and $T_{max}$, maximum no. of iterations

\begin{algorithmic}[1] %[1] enables line 
\STATE Initialize training data of function evaluations $\mathcal{D}$
\STATE Initialize statistical models $\mathcal{M}_1, \cdots, \mathcal{M}_k,,\mathcal{M}_{C_0},\cdots, \mathcal{M}_{C_{m}}$  from $\mathcal{D}$
\FOR{each iteration $t$=1 to $T_{max}$}
\STATE // Solve constrained cheap MO problem  with objectives \textsc{Af}$(\mathcal{M}_1,x),\cdots, \textsc{Af}(\mathcal{M}_k,x)$ to get candidate inputs
\STATE $\mathcal{X}_{p} \leftarrow \arg min_{x \in \mathcal{X}} (\textsc{Af}(\mathcal{M}_1,x),\cdots, \textsc{Af}(\mathcal{M}_k,x))$ \\ \qquad \qquad \textbf{s.t}  $\mu_{c_0},\cdots,\mu_{c_m},C_{m+1}\cdots, C_L$

\STATE // Pick the candidate input with maximum uncertainty
\STATE Select $x_{t+1} \leftarrow \arg max_{x\in \mathcal{X}_{p}} \; U_{\beta_t}(x)$
\STATE Evaluate $x_{t+1}$: $Y_{t+1} \leftarrow (F_1(x_{t+1}),\cdots,F_k(x_{t+1}))$, $C_{t+1} \leftarrow (C_0(x_{t+1}),\cdots,C_L(x_{t+1}))$ %$y_i=f_i(x_{t+1}) ~ i=1,\dots,K$
\STATE Aggregate data: $\mathcal{D} \leftarrow \mathcal{D} \cup \{(x_{t+1}, Y_{t+1}, C_{t+1})\}$ %$D^{(t+1)}=D^{(t)} \cup {(x_{t+1},(y_1,\dots,y_K))}$
\STATE Update models $\mathcal{M}_1,\cdots, \mathcal{M}_k,\mathcal{M}_{C_0},\cdots, \mathcal{M}_{C_{m}}$  using $\mathcal{D}$ %$M_i^{(t+1)} \leftarrow $ posterior of $M_i(D^{(t+1)})  ~ i=1,\dots,K$
\STATE $t \leftarrow t+1$
\ENDFOR
\STATE \textbf{return} Pareto set and Pareto front of $\mathcal{D}$
\end{algorithmic}
%\vspace{-1.0ex}
\end{algorithm}
\vspace{0.8ex}

\noindent {\bf Advantages.} USeMOC has many advantages over prior methods. {\bf 1)} Provides flexibility to plug-in any acquisition function for single-objective BO. This allows us to leverage existing acquisition functions including EI and LCB. {\bf 2)} Computationally-efficient to solve constrained MO problems with many objectives. {\bf 3)} Can handle all three types of constraints (i.e., type 1, type 2, and type 3).

%\vspace{-0.3cm}
\subsection{Key Algorithmic Components of USeMOC} 
The two main algorithmic components of USeMOC framework are: selecting most promising candidate inputs by solving a cheap constrained MO problem and picking the best candidate via uncertainty maximization. We describe their details below.

\vspace{0.8ex}

\noindent {\bf Selection of promising candidates.} We employ the statistical models $\mathcal{M}_1, \mathcal{M}_2,\cdots, \mathcal{M}_k$ towards the goal of selecting promising candidate inputs as follows. Given an acquisition function \textsc{Af} (e.g., EI), we construct a constrained cheap multi-objective optimization problem with objectives $\textsc{Af}(\mathcal{M}_1,x), \textsc{Af}(\mathcal{M}_2,x),\cdots, \textsc{Af}(\mathcal{M}_k,x)$, where $\mathcal{M}_i$ is the statistical model for unknown function $F_i$ and  $C_1, C_2,\cdots, C_L$ are the problem constraints. Generally, there are three types of constraints: 

\begin{itemize}
    \item \textbf{[Type 1]} $C_i$ is a function of the input $x$ and can be expressed declaratively (white-box constraint). Such a constraint will take the same form as in the cheap MO problem.
    
    \item \textbf{[Type 2]} $C_i$ is black-box constraint. In this case, it will be modeled by an independent GP $\mathcal{M}_{c_i}$ and predictive mean of its model $\mu_{c_i}$ can be used in the optimization process.
    
    \item  \textbf{[Type 3]} $C_i$ is a function of both the input $x$ and a combination of the black-box objective functions. Since verifying constraint $C_i$ depends on the evaluation of one (or more) objectives $F_j$, we employ the predictive mean(s) $\mu_i(x)$ instead.
\end{itemize}
Without loss of generality and for the sake of notation in Algorithm \ref{alg:USeMO}, we suppose that the first $m$ constraints with $0\leq m \leq L$ are black-box.

Since we present the framework as minimization, all AFs will be minimized. The Pareto set $\mathcal{X}_p$ obtained by solving this cheap constrained MO problem represents the most promising candidate inputs for evaluation.
%\vspace{-0.2cm}
\begin{align}
% \scriptsize
\label{eq2}
\mathcal{X}_p\leftarrow \arg min_{x \in \mathcal{X}} \; &(\textsc{Af}(\mathcal{M}_1,x),\cdots, \textsc{Af}(\mathcal{M}_k,x))\\
&  \textbf{s.t } \mu_{c_0},\cdots,\mu_{c_m},C_{m+1}\cdots, C_L \nonumber
% \vspace{-0.5cm}
\end{align}
Each acquisition function \textsc{Af}($\mathcal{M}_i,x$) is dependent on the corresponding surrogate model $\mathcal{M}_i$ of the unknown objective function $F_i$. Hence, each acquisition function will carry the information of its associated objective function. As iterations progress, using more training data, the models $\mathcal{M}_1, \mathcal{M}_2,\cdots, \mathcal{M}_k$ will better mimic the true objective functions $F_1, F_2,\cdots,F_k$. Therefore, the Pareto set of the acquisition function space (solution of Equation~\ref{eq2}) becomes closer to the Pareto set of the true functions $\mathcal{X}^*$ with increasing iterations.

Intuitively, the acquisition function \textsc{Af}($\mathcal{M}_i,x$) corresponding to unknown objective function $F_i$ tells us the utility of a point $x$ for optimizing $F_i$. The input minimizing \textsc{Af}($\mathcal{M}_i,x$) has the highest utility for $F_i$, but may have a lower utility for a different function $F_j$ ($j \neq i$). The utility of inputs for evaluation of $F_j$ is captured by its own acquisition function \textsc{Af}($\mathcal{M}_j,x$). Therefore, there is a trade-off in the utility space for all $k$ different functions. The Pareto set $\mathcal{X}_p$ obtained by simultaneously optimizing acquisition functions for all $k$ unknown functions will capture this utility trade-off. As a result, each input $x \in \mathcal{X}_p$ is a promising candidate for evaluation towards the goal of solving MOO problem. 
USeMOC employs the same acquisition function for all $k$ objectives. The main reason is to give equivalent evaluation for all functions in the Pareto front (PF) at each iteration. If we use different AFs for different objectives, the sampling procedure would be different. Additionally, the values of various AFs can have considerably different ranges. Thus, this can result in an unbalanced trade-off between functions in the cheap PF leading to the same unbalance in our final PF.

\vspace{1.0ex}

\noindent{\bf Cheap constrained MO solver.} We employ the constrained version of the popular NSGA-II algorithm (\cite{deb2002nsga,deb2002fast}) to solve the MO problem with cheap objective functions and cheap constraints noting that any other algorithm can be used.

\vspace{1.0ex}

\noindent {\bf Picking the best candidate input.} We need to select the best input from the Pareto set $\mathcal{X}_p$ obtained by solving the cheap MO problem. All inputs in $\mathcal{X}_p$ are promising in the sense that they represent the trade-offs in the utility space corresponding to different unknown functions. It is critical to select the input that will guide the overall search towards the goal of quickly approximating the true Pareto set $\mathcal{X}^*$. We employ a uncertainty measure defined in terms of the statistical models $\mathcal{M}_1, \mathcal{M}_2,\cdots, \mathcal{M}_k$ to select the most promising candidate input for evaluation. In single-objective optimization case, the learned model's uncertainty for an input can be defined in terms of the variance of the statistical model. For multi-objective optimization case, we define the uncertainty measure as the volume of the uncertainty hyper-rectangle.
%\vspace{-0.2cm}
\begin{align}
\label{eq1}
U_{\beta_t}(x) =& VOL(  \{(LCB(\mathcal{M}_i, x), UCB(\mathcal{M}_i, x) \}_{i=1}^{k} )
\end{align}
where LCB$(\mathcal{M}_i, x)$ and UCB$(\mathcal{M}_i, x)$ represent the lower confidence bound and upper confidence bound of the statistical model $\mathcal{M}_i$ for an input $x$ as defined in equations ~\ref{ucbeq} and \ref{lcbeq}; and $\beta_t$ is the parameter value to trade-off exploitation and exploration at iteration $t$. We employ the adaptive rate recommended by (\cite{gp-ucb}) to set the $\beta_t$ value depending on the iteration number $t$. We measure the uncertainty volume measure for all inputs $x \in \mathcal{X}_p$ and select the input with maximum uncertainty for function evaluation.
%\vspace{-0.2cm}
\begin{align}
    x_{t+1}=\arg max_{x\in \mathcal{X}_p} \; U_{\beta_t}(x)
\end{align}
% \vspace{-0.5cm}
\section{Experiments and Results}
In this section, we discuss the experimental evaluation of USeMOC and prior methods on a real-world analog circuit design optimization task. The code for USeMOC is available in github repository: github.com/belakaria/USEMOC.
%\vspace{-0.5cm}
%\subsection{Experimental Setup}

\vspace{1.0ex}

\noindent {\bf Analog circuit optimization domain.} We consider optimizing the design of a multi-output switched-capacitor voltage regulator via Cadence circuit simulator that imitates the real hardware \cite{DATE-2020}. Each candidate circuit design is defined by 32 input variables ($d$=32). The first 24 variables are the width, length, and unit of the eight capacitors of the circuit $W_i,L_i,M_i ~ \forall i \in 1\cdots 8$. The remaining input variables are four output voltage references $V_{ref_i}~ \forall i \in 1\cdots 4$ and four resistances $R_i ~ \forall i \in 1\cdots 4$. We optimize nine objectives: maximize efficiency $Eff$, maximize four output voltages $V_{o_1} \cdots V_{o_4}$, and minimize four output ripples $OR_{1} \cdots OR_{4}$.
Our problem has a total of nine constraints:
%\vspace{-0.5cm}
\begin{align}
    &{C}_0: Cp_{total} \simeq 20 nF  ~ with ~ Cp_{total}= \sum_{i=1}^8 (1.955W_iL_i+0.54(W_i+L_i))M_i \nonumber\\
    &{C}_1~ to~ {C}_4:  V_{o_i}\geq V_{ref_i} ~ \forall \in{1\cdots4} \nonumber\\
    &{C}_5\ to\ {C}_{8\ }:\ \ OR_{lb}\le OR_i\le OR_{ub} ~ \forall i\ \in{1\cdots4} \nonumber\\
    &{C}_9:Eff \le 100\%\nonumber
\end{align}
where $OR_{lb}$ and $OR_{ub}$ are the predefined lower-bound and upper-bound of $OR_i$ respectively. $Cp_{total}$ is the total capacitance of the circuit. In this problem, $C_0$ is a white-box constraint (Type 1), while remaining constraints are combinations of black-box objectives (Type 3).

\vspace{0.8ex}

\noindent {\bf Multi-objective BO algorithms.} We compare USeMOC with the existing BO method PESMOC. %We present the results of USeMOC with EI and LCB acquisition functions: USeMOC-EI and USeMOC-LCB. 
Due to lack of BO approaches for constrained MO, we compare to known genetic algorithms (NSGA-II and MOEAD). However, they require large number of function evaluations to converge which is not practical for optimization of expensive functions. 

\vspace{0.8ex}

\noindent {\bf Evaluation Metrics.} To measure the performance of baselines and USeMOC, we employ two different metrics, one measuring the accuracy of solutions and another one measuring the efficiency in terms of the number of simulations.  1) Pareto hypervolume (PHV) is a commonly employed metric to measure the quality of a given Pareto front \cite{zitzler1999evolutionary}. %PHV is defined as the volume between a reference point and the given Pareto front. 
After each iteration $t$ (or number of simulations), we measure the PHV for all algorithms. We evaluate all algorithms for 100 circuit simulations. 2) Percentage gain in simulations is the fraction of simulations our BO algorithm USeMOC is saving to reach the PHV accuracy of solutions at the convergence point of baseline algorithm employed for comparison. 

\vspace{1.0ex}

\begin{figure} [h!]
\centering
% \begin{subfigure}{.45\textwidth}
%   \centering
  \includegraphics[width=0.5\linewidth]{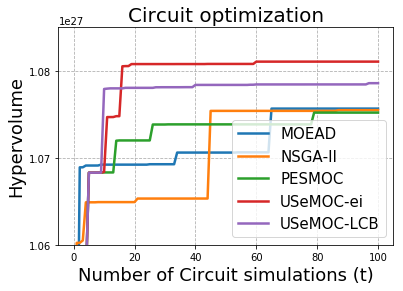}
%   \label{fig:test1}
\caption{Results of different multi-objective algorithms including USeMOC. The hypervolume metric is shown as a function of the number of circuit design simulations.}\label{fig:hv}
% \end{subfigure}%

\end{figure}

\noindent {\bf Results and Discussion.} We evaluate the performance of USeMOC with two different acquisition functions (EI and LCB) to show the generality and robustness of our approach. We also provide results for the percentage gain in simulations achieved by USeMOC when compared to each baseline method. Figure \ref{fig:hv} shows the PHV metric achieved by different multi-objective methods including USeMOC as a function of the number of circuit simulations. We make the following observations: 1) USeMOC with both EI and LCB acquisition functions perform significantly better than all baseline methods. 2) USeMOC is able to uncover a better Pareto solutions than baselines using significantly less number of circuit simulations. This result shows the efficiency of our approach. Table \ref{tab:costreduction} shows that USeMOC achieves percentage gain in simulations w.r.t baseline methods ranging from 90 to 93\%.\\

\begin{figure}[h!]
  \centering
  \includegraphics[width=.55\linewidth]{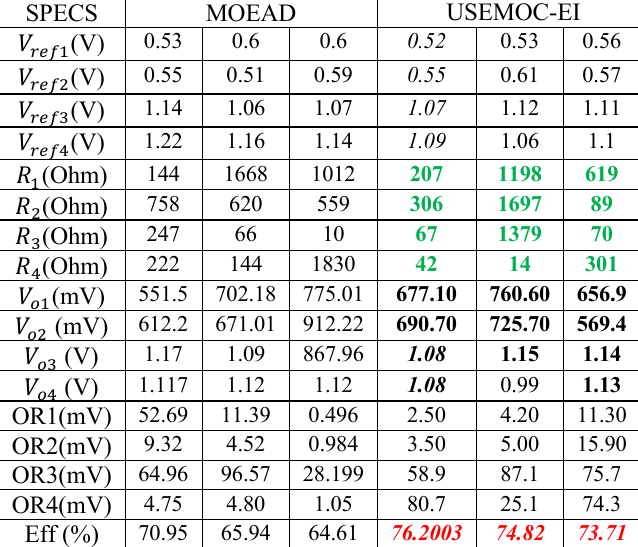}

\caption{Comparison table of optimized circuit parameters obtained by MOEAD and USeMOC-EI (designs are selected from the Pareto set prioritized by efficiency)}  \label{fig:table}
\end{figure}

\noindent The analog circuit is implemented in the industry-provided process design kit (PDK) and shows better efficiency and output ripples. Since MOEAD is the best performing baseline optimization method, we use it for the rest of the experimental analysis. Table \ref{fig:table} illustrates the simulated performance of circuit optimized by MOEAD (best baseline) and USeMOC-EI (best variant of our proposed algorithm). Results of both algorithms meet the voltage reference and ripple requirements (100mV) . Compared to MOEAD, the optimized circuit with USeMOC-EI can achieve a higher conversion efficiency of 76.2 \% (5.25 \% higher than MOEAD, highlighted in red color) with similar output ripples.
The optimized circuits can generate the target output voltages within the range of 0.52V-0.61V (1/3x ratio) and 1.07V-1.12V (2/3x ratio) under the loads varying from 14 Ohms to 1697 Ohms (highlighted in black and green colors). Thus, the capability of USeMOC to optimize the parameters of circuit under different output voltage/current conditions is clearly validated. Future work includes improving USeMOC to solve more challenging problems \cite{belakaria2020PSD}.

 \begin{table}[ht!]
% \vspace{-1.0ex}
\centering
\resizebox{0.6\linewidth}{!}{
\begin{tabular}{lllllll}  
\toprule
Method & MOEAD  & NSGA-II &  PESMOC  \\
\midrule
Gain in simulations & 90.7\%  & 93.3\%  &92.5\%  \\
\bottomrule
\end{tabular}}
\caption{Percentage gain in simulations achieved by our USeMOC  compared to baselines.}
\label{tab:costreduction}
\end{table} 

%\bibliography{ref}

% Appendix goes to a new page

% \newpage

% \appendix
% \input{appendix}

\end{document}